\documentclass[10pt,twocolumn,letterpaper]{article}

\usepackage{cvpr}              

\usepackage{graphicx}
\usepackage{amsmath}
\usepackage{amssymb}
\usepackage{booktabs}
\usepackage{multirow}

\usepackage[accsupp]{axessibility}

\newcommand{\tabincell}[2]{\begin{tabular}{@{}#1@{}}#2\end{tabular}}
\usepackage[pagebackref,breaklinks,colorlinks]{hyperref}

\usepackage[capitalize]{cleveref}
\crefname{section}{Sec.}{Secs.}
\Crefname{section}{Section}{Sections}
\Crefname{table}{Table}{Tables}
\crefname{table}{Tab.}{Tabs.}

\def\cvprPaperID{106} 
\def\confName{CVPRW}
\def\confYear{2022}

\usepackage{url}

\def \ie{{\em i.e.}}

\usepackage{hyperref}
\hypersetup{
	colorlinks=true,
	linkcolor=blue,
	filecolor=blue,      
	urlcolor=blue,
	citecolor=cyan,
}

\begin{document}

\title{Progressive Training of A Two-Stage Framework for Video Restoration}

\author{Meisong~Zheng\textsuperscript{1}\footnotemark[1], Qunliang~Xing\textsuperscript{1}\footnotemark[1], Minglang~Qiao\textsuperscript{1}\footnotemark[1], Mai~Xu\footnotemark[2], Lai~Jiang, Huaida~Liu\textsuperscript{1} and Ying~Chen\textsuperscript{1}\footnotemark[2]\\
\textsuperscript{1}Alibaba Group\\
{\tt\small \{zhengmeisong.zms, xingqunliang.xql, qiaominglang.qml\}@alibaba-inc.com}\\
{\tt\small xumai@icloud.com, jianglai.china@gmail.com, \{liuhuaida.lhd, yingchen\}@alibaba-inc.com}
}

\thispagestyle{empty}
\twocolumn[{%
    \renewcommand\twocolumn[1][]{#1}%
    \vspace{-1em}
    \maketitle
    \vspace{-1em}
    \begin{center}
        \centering
        \vspace{-0.5cm}
        \includegraphics[width=0.99\textwidth]{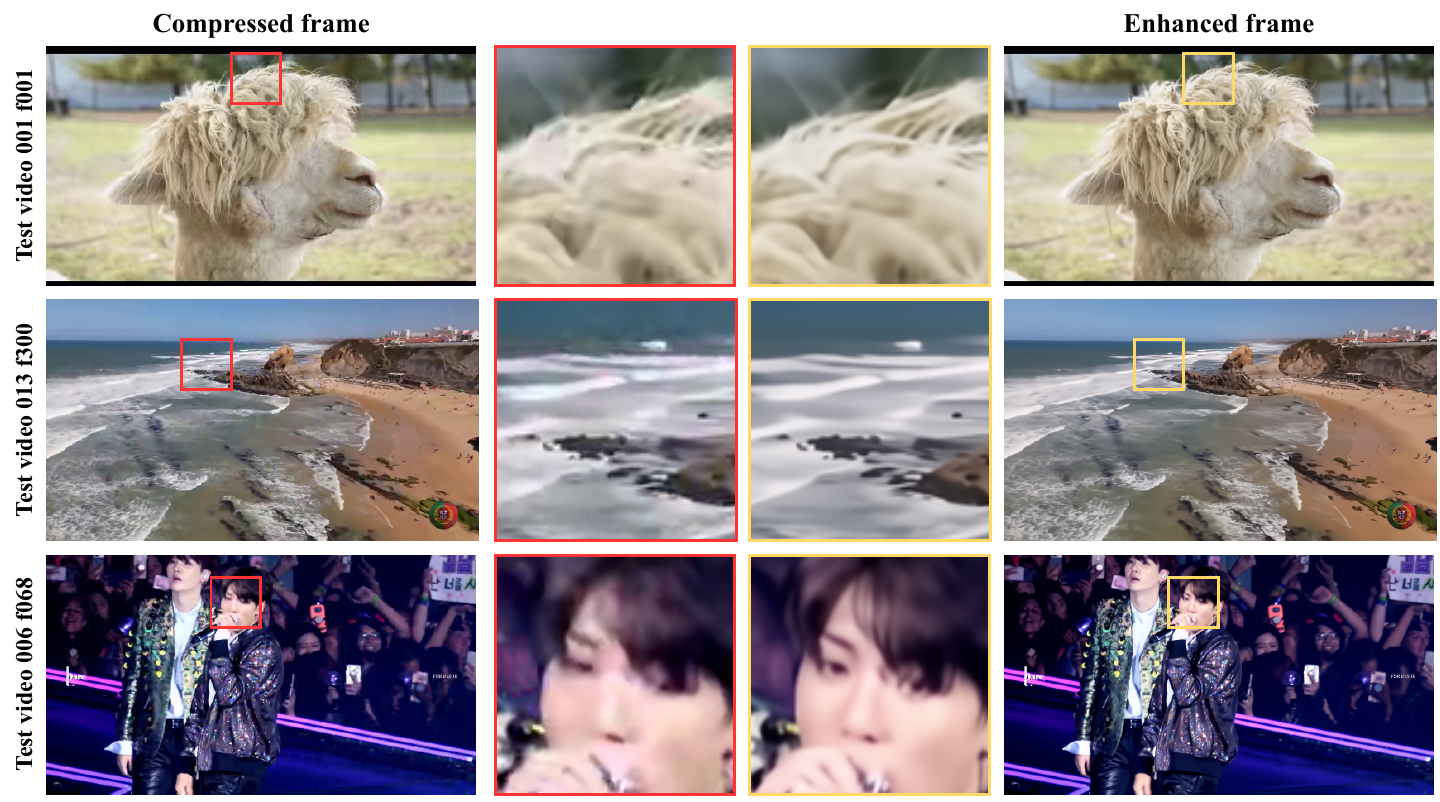}\vspace{0.1cm}
        \vskip -0.2cm
        \captionof{figure}{Subjective performance of our proposed method on the test set of the NTIRE 2022 challenge.}
        \label{fig:sub1}
        \vspace{0.15cm}
    \end{center}%
}]

\begin{abstract}
As a widely studied task, video restoration aims to enhance the quality of the videos with multiple potential degradations, such as noises, blurs and compression artifacts. Among video restorations, compressed video quality enhancement and video super-resolution are two of the main tacks with significant values in practical scenarios. Recently, recurrent neural networks and transformers attract increasing research interests in this field, due to their impressive capability in sequence-to-sequence modeling. However, the training of these models is not only costly but also relatively hard to converge, with gradient exploding and vanishing problems. To cope with these problems, we proposed a two-stage framework including a multi-frame recurrent network and a single-frame transformer. Besides, multiple training strategies, such as transfer learning and progressive training, are developed to shorten the training time and improve the model performance. Benefiting from the above technical contributions, our solution wins two champions and a runner-up in the NTIRE 2022 super-resolution and quality enhancement of compressed video challenges.
Code is available at \url{https://github.com/ryanxingql/winner-ntire22-vqe}.
\footnotetext[1]{These authors contributed equally to this work.}
\footnotetext[2]{Corresponding authors.}
\end{abstract}

\section{Introduction}
\label{sec:intro}

The recent decades have witnessed an explosive growth of video data over the internet.
Meanwhile, the resolution of the videos becomes higher and higher to satisfy the increasing demand for the quality of experience (QoE). However, due to the limited bandwidth, the videos are commonly down-sampled and compressed, which causes inevitably degradation on video quality. Therefore,  
it draws a great attention in the computer vision community for video restoration tasks, such as video super-resolution, de-artifacts of compressed video.

Video restoration is challenging because it requires aggregating information from multiple highly related but misaligned low-quality frames in video sequences.
Most existing methods of video restoration consider it as a spatial-temporal sequence prediction problem, and can be mainly divided into two categories: sliding window methods~\cite{wang2019edvr,yang2018multi,guan2019mfqe,Deng_Wang_Pu_Zhuo_2020stdf,xing2020early} and recurrent-based methods~\cite{kelvin2021Basicvsrpp,baidu2022ppmvsr,kelvin2020basicvsr}.
For instance, BasicVSR++~\cite{kelvin2021Basicvsrpp} proposes a second-order grid propagation network to better mining the spatial-temporal information.
It demonstrates the great effectiveness of the recurrent framework and wins the NTIRE 2021 quality enhancement of heavily compressed video challenge.
However, the recurrent framework processes the video frames sequentially, which limits the efficiency of the recurrent-based methods. 
Recent works~\cite{cao2021vsrt,liang2022vrt} try to enhance video frames in parallel, on the top of transformer architecture. However, both recurrent network and transformer have square computational complexity with respect to sequence length and image size, resulting in O(n4) computational complexity.
Subject to the huge memory consumption, these networks can only be fed by the clipped sequence with no more than 16 frames, even on a NVIDIA A100 GPU. This degrades PSNR performance compared to BasicVSR++\cite{kelvin2021Basicvsrpp} on the REDs dataset~\cite{Nah_2019_CVPR_Workshops_REDS}.
Besides the large consumption of GPU memory, the models with larger structures, such as Transformer, are also hard to be tuned. That is, we sometimes are unable to finely adjust the key hyper-parameters, like batch size and learning rate, which are essential on stabilizing the training process. 
Moreover, the ``large'' models also prone to suffer from the problems of over-fitting and performance fluctuation across the restored frames.

To address the above problems, we propose a two-stage framework combing a multi-frame recurrent-based network and single-frame transformer-based network. Specifically, the first stage is developed to coarsely restore the video frames and alleviate the quality fluctuation across the frames. Given the restored frames from the first stage, the second stage further effectively removes the severe artifacts frame by frame. 
Specifically, the first stage model is an improved BasicVSR++\cite{kelvin2021Basicvsrpp}, and in the second stage we adopt SwinIR\cite{liang2021swinir} as the backbone model.
We train these two models separately to save memory resources and further improve the accuracy.
Besides, multiple strategies of transfer learning and progressive training are conducted in both two stages, to not only accelerate the convergence but also improve final restoration performance.
In summary, the contributions of this paper are as follows:

\begin{itemize}
    \item {We propose a two-stage framework to simultaneously remove compression artifacts and mitigate the quality fluctuation in compressed videos.}
    \item {We introduce a progressive training scheme to stabilize training and improve finally performance.}
    \item {We introduce a transfer learning strategy with pre-trained models to shorten training time.}
    \item {Our proposed method achieves a good trade-off between the enhancement performance and model complexity, and wins the NTIRE 2022 challenge of super-resolution and quality enhancement of compressed video~\cite{yang2022ntire}.}
\end{itemize}

\section{Related Work}
\label{sec:formatting}

\subsection{Video Restoration}

As one of the main tracks of video restoration, compressed video quality enhancement on has been widely studied~\cite{yang2017dscnn, dong2015arcnn, xing2020early, yang2018multi, guan2019mfqe} in the past years. 
Among them, most of the existing methods are based on the single-frame quality enhancement~\cite{yang2017dscnn, dong2015arcnn, xing2020early}. 
Observing that the frame quality remarkably fluctuates after compression, MFQE~\cite{yang2018multi} and its extended version MFQE 2.0~\cite{guan2019mfqe} take advantage of neighboring high-quality frames.
They adopt a temporal fusion scheme that incorporates dense optical flow for motion compensation.
Similarly, STDF~\cite{Deng_Wang_Pu_Zhuo_2020stdf} aggregates temporal information while avoiding explicit optical flow estimation.
  
\textbf{Video Super Resolution.}
In addition to the compressed video quality enhancement, video super resolution (VSR) aims to restore the videos by improving their resolution. 
Different from the single image super resolution (SISR), VSR utilizes neighboring frames to reconstruct the high-resolution sequence.
The existing VSR methods can be divided into two categories: window-based~\cite{li2020mucan, xue2017tof, wang2019edvr, D3Dnet} and recurrent methods~\cite{kelvin2020basicvsr, kelvin2021Basicvsrpp, takashi2020rsdn, baidu2022ppmvsr}.
Specifically, EDVR~\cite{wang2019edvr} adopts deformable convolutions~\cite{dai2017dcn,zhu2018dcnv2} to align neighbouring frames.
Similar to EDVR, D3DNet~\cite{D3Dnet} uses deformable 3D convolution network to fully exploit the spatio-temporal information for video SR.
Besides, BasicVSR~\cite{kelvin2020basicvsr} proposes to untangle the basic components for VSR such as propagation, alignment, aggregation and up-sampling. 
On the top of BasicVSR, BasicVSR++~\cite{kelvin2021Basicvsrpp} further improves performance with extensive bi-directional propagation strategy and flow-guided deformable alignment.
In this work, we adopt BasicVSR++ as our backbone model in the first stage.

\subsection{Vision Transformer}

\begin{figure*}[htb]
  \centering
  \includegraphics[width=0.8\linewidth]{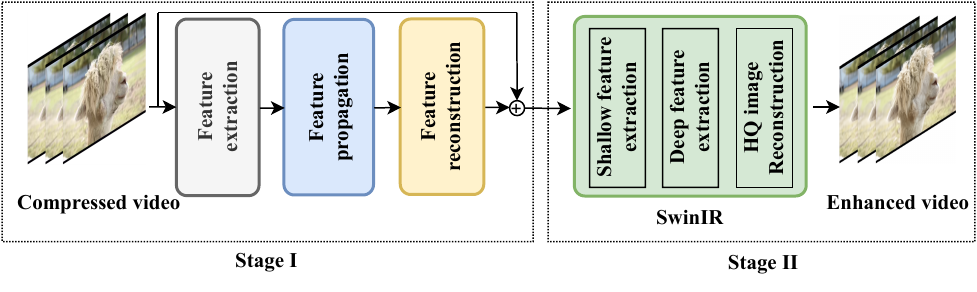}
   \caption{Two-stage framework of our method.}
   \label{fig:framework}
\end{figure*}

Recently, sourced from the area of natural language processing (NLP)~\cite{BERT2018,albert2019,roberta2019}, Transformers have shown the outstanding performance and outperforms the  state-of-the-art models in many vision tasks, including image classification, object detection, semantic segmentation, human pose estimation and video classification~\cite{vit2020,swin2021,mae2021,detr2020,swin2021,hrformer2021,xia2022vision,arnab2021vivit,xia2022vision}. Specifically, Swin transformer~\cite{swin2021} proposes a hierarchical transformer structure with shifted windows mechanism, which integrates the advantages of build-in inductive biases of CNN and long-range self-attention of transformers.

There also exists some attempts to apply transformers in low-level vision tasks~\cite{chen2020ipt,liang2021swinir,uformer2021,restormer2021,yang2020learning,iqatransformer2021,jiang2021transgan}. For instance, SwinIR~\cite{liang2021swinir} proposes an image restoration model based on Swin transformer, which not only handles local context but also efficiently captures long-range dependencies.
Uformer~\cite{uformer2021} proposes a general U-shaped transformer-based structure, which shows strong performance on real de-noising tasks.

Transformers have also been introduced for video restoration~\cite{cao2021vsrt,liang2022vrt,fuoli2022dap}.
VSRT~\cite{cao2021vsrt} utilizes the parallel computing ability of transformer to align the features between neighboring frames in parallel.
VRT\cite{liang2022vrt} introduces a temporal mutual self attention module to better mining spatial-temporal information.
Unfortunately, these approaches can not be trained with longer video clips as they require large memory of GPU.
In this work, we adopt SwinIR as our backbone model in the second stage.

\section{Method}
\label{sec:formatting}

\subsection{Proposed Two-stage Framework}

\begin{figure*}[htb]
  \centering
  \includegraphics[width=0.9\linewidth]{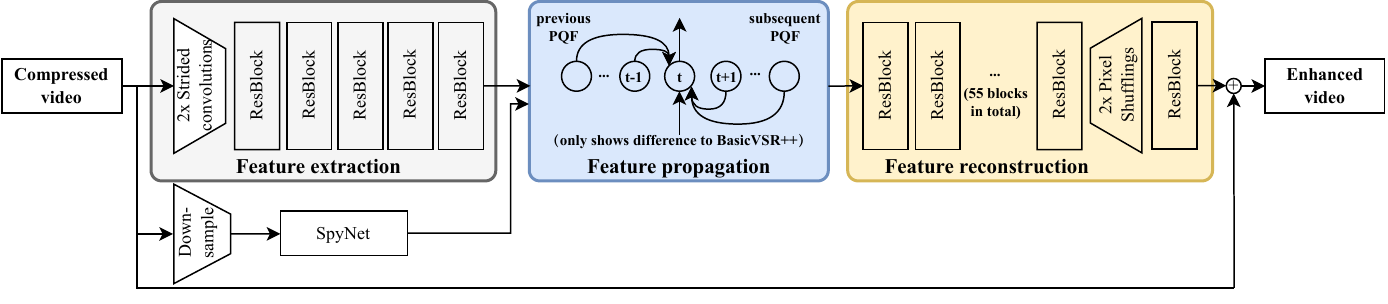}
   \caption{Structure of stage I model (for Track 1).}
   \label{fig:stage1}
\end{figure*}

We first introduce our two-stage framework for video restoration, as shown in Fig.~\ref{fig:framework}.
In stage I, the network is developed on the top of BasicVSR++~\cite{kelvin2021Basicvsrpp}.
Based on this, we replace the second-order flows in BasicVSR++ by PQF flows~\cite{yang2018multi,guan2019mfqe}.
Besides, we deepen the reconstruction module of BasicVSR++ from 5 residual blocks to 55 blocks.
In stage II, we further improve the quality of the enhanced consecutive frames by a state-of-the-art image restoration network, \ie, SwinIR~\cite{liang2021swinir}.
This stage helps remove severe artifacts and further improve the quality upon the previous stage.
Finally, the networks of stage I and II are cascaded for producing the final results.
In summary, we first feed the compressed video with $N$ compressed frames $\{F_t\}_{t=1}^N$ into the stage I model.
Then, we obtain the enhanced video frames $\{\tilde{F_t}\}_{t=1}^N$ by stage I.
Next, we feed $\{\tilde{F_t}\}_{t=1}^N$ into the stage II model frame by frame.
Finally, we get the enhanced video frames $\{\hat{F_t}\}_{t=1}^N$, which are sequentially combined into the final enhanced video.

\subsection{First Stage and Progressive Training}

Our stage I model consists of three developed modules: feature extraction, propagation and image reconstruction.
Given an input video, two strided convolution and five residual blocks are first applied to extract spatial features from the input frames.
At the same time, all input frames are down-sampled by the factor of 4 with an bicubic filter, and then applied to SpyNet~\cite{ranjan2017optical} to calculate the forward and backward flows.
Next, as shown in Fig.~\ref{fig:stage1}, for enhancing the $t$-th frame, the features of neighboring $(t-1)$-th and $(t+1)$-th frames as well as the features of previous and subsequent PQFs are propagated to the spatial feature of the $t$-th frame.
For the propagation of each frame, the frame is warped by its estimated flow.
Finally, we use 55 residual blocks to decode the propagated features, and reconstruct the video. To be more specific, in stage I model, 
we use pixel shuffling~\cite{shi2016real} to restore the resolution of decoded features. Besides, 
residual learning~\cite{he2016deep} is also conducted for generating the final enhanced image, by reducing the training complexity of the model.

As introduced, our reconstruction module contains 55 residual blocks, which is rather ``heavy'' for training. Thus, a progressive training~\cite{ntire2020vmp,peng2020mwcnn} strategy is conducted for our stage I model.
Specially, we lighten the reconstruction module by using its first 5, 15, 25, 35, 45 and 55 residual blocks for reconstruction, respectively. Specifically, let $R_1$, $R_2$, ..., $R_5$ and $R_6$ denote the 1-5, 6-15, 16-25, 26-35, 36-45, 46-55 residual blocks; $E$ and $P$ refer to the modules of feature extraction and feature propagation; $S$ and $R$ are the two pixel shuffling layers and residual block at the end of our stage I model. Given the input frame $I_{\text{in}}$, the restored frame $I_{\text{out}}$ can be obtained by the progressively training as follows
\begin{align}
    I_{\text{out}} &= R(S(R_1(P(E(I_{\text{in}})))))\\
    I_{\text{out}} &= R(S(R_2(R_1(P(E(I_{\text{in}}))))))\\
    I_{\text{out}} &= R(S(R_3(R_2(R_1(P(E(I_{\text{in}})))))))\\
    I_{\text{out}} &= R(S(R_4(R_3(R_2(R_1(P(E(I_{\text{in}}))))))))\\
    I_{\text{out}} &= R(S(R_5(R_4(R_3(R_2(R_1(P(E(I_{\text{in}})))))))))\\
    I_{\text{out}} &= R(S(R_6(R_5(R_4(R_3(R_2(R_1(P(E(I_{\text{in}}))))))))))
\end{align}
For the first training, we load the parameters of $E$, $P$, $S$ and $R$ from the open-sourced model of BasicVSR++.
For the k-th training ($2<=k<=6$), we load the parameters of $E$, $P$, $S$, $R$ and $\{R_i\}_{i=1}^{k-1}$ from the $(k-1)$-th converged model. Note that the temporal information is embedded in the propagation module as illustrated in Fig.~\ref{fig:framework}, which is simplified in the above equations.

\subsection{Second Stage and Transfer Learning}
Although a single BasicVSR++ could achieve state-of-the-art performance for the compressed videos restoration, the restored results are not satisfactory in the cases with severely distorted scenes. 
Thus, we develop a stage II model to further refine the enhanced video frames by stage I model, similar to the two-stage restoration strategy in \cite{wang2019edvr}.
However, different from \cite{wang2019edvr}, we empirically find that simply cascading a second BasicVSR++ on stage I can only bring slight improvement.
Instead of cascading a video restoration model, we employ a single-image restoration model in stage II to further improve the quality of the enhanced frames.

Specifically, SwinIR \cite{liang2021swinir} model is utilized in stage II to further enhance the outputs of stage I, which is proven to be still effective for enhancement of compressed video, in addition to the restoration of single image.
Besides, due to the fact that transformer requires training in large-scale datasets, transfer learning is applied during the training of SwinIR.
More specifically, the SwinIR model is initialized by pre-trained parameters from \cite{liang2021swinir}, which is trained for RGB image denoising.
The effectiveness of stage II is illustrated in Table \ref{tab:tenvideo}.

\begin{table*}[htbp]
  \centering
  \caption{Quantitative results of average PSNR $\uparrow$ on stage I model in Track 1. Note that LDV refers to the 230 official training data, and EX refers to the 870 training data collected from YouTube. MSE is the Mean Squared Error training loss, and RMD indicates that we removed duplicated frames at test time.}
  \resizebox{0.7 \textwidth}{!}{
    \begin{tabular}{c|cccc}
    \toprule
    Model & Params & Settings & Our Offline & Official(10th frames) \\
    \midrule
    BasicVSR++\_c128n25\cite{kelvin2021Basicvsrpp} & 44.08M & LDV     & 32.5930 & 31.8269 \\
    StageI\_c128n25 & 44.08M & LDV  & 32.6130 & 31.8611 \\
    \hline
    StageI\_c128n25 & 44.08M & LDV+EX    & 32.6957 & 32.0252 \\
    StageI\_c128n25\_rec2 & 47.51M & LDV+EX    & 32.7484 & 32.0587 \\
    StageI\_c128n25\_rec3 & 50.61M & LDV+EX    & 32.7850 & 32.0826 \\
    StageI\_c128n25\_rec4 & 53.71M & LDV+EX    & 32.7945 & 32.0933 \\
    StageI\_c128n25\_rec5 & 56.80M & LDV+EX    & 32.8054 & 32.1025 \\
    StageI\_c128n25\_rec6 & 59.90M & LDV+EX    & 32.8240 & 32.1067 \\
    \hline
    StageI\_c128n25\_rec6 & 59.90M & LDV+cleaned\_EX & 32.8672 & 32.1934 \\
    StageI\_c128n25\_rec6 & 59.90M & LDV+cleaned\_EX, MSE & 32.8968 & 32.2224 \\
    StageI\_c128n25\_rec6 & 59.90M & LDV, MSE & 32.9055 & 32.2323 \\
    StageI\_c128n25\_rec6 & 59.90M & LDV, MSE, RMD & 32.9193 & 32.2395 \\
    \bottomrule
    \end{tabular}%
    }
  \label{tab:all}%
\end{table*}%

\begin{table*}[htbp]
  \centering
  \caption{Quantitative results of PSNR $\uparrow$ on our 10 offline validation videos in Track 1. Note that LDV refers to the 230 official training data, and EX refers to the 870 training data collected from YouTube. MSE is the Mean Squared Error training loss; RMD indicates that we removed duplicated frames at test time, and TTA indicates the employing of self-ensemble. {TTA\_I and TTA\_II indicates applying self-ensemble in stage I and II, respectively.}}
  \resizebox{0.99 \textwidth}{!}{
    \begin{tabular}{c|c|c|ccccccccccc|c}
    \toprule
    Stage & Model & Settings & 30    & 56    & 102   & 106   & 109   & 119   & 124   & 125   & 158   & 189   & \tabincell{c}{Avg.\\Offline} & \tabincell{c}{Avg.\\Offical} \\
    \hline
        & LQ Input & - & 29.39 & 32.89 & 27.84 & 34.09 & 30.04 & 28.90  & 30.14 & 34.19 & 31.9  & 26.79 & 30.6170 & 30.1768 \\
    \hline
    \multirow{2}[2]{*}{Baseline} & \tabincell{c}{BasicVSR++\_c128n25\cite{kelvin2021Basicvsrpp}} & LDV & 31.65 & 35.13 & 29.08 & 35.65 & 31.00 & 30.30 & 32.84 & 37.40  & 34.75 & 28.14 & 32.5930 & 31.8269 \\
          & \tabincell{c}{BasicVSR++\_c128n25\cite{kelvin2021Basicvsrpp}} & LDV, TTA & 31.90 & 35.29 & 29.20 & 35.70 & 31.30 & 30.42 & 33.19 & 37.71  & 35.00 & 28.30 & 32.8019 & 32.1188 \\
    \hline
    \multirow{7}[2]{*}{I} & Stage I\_c128n25 & LDV & 31.64 & 35.01 & 29.07 & 25.64 & 31.17 & 30.29 & 32.88 & 37.47 & 34.82 & 28.13 & 32.6130 & 31.8611\\
          & StageI\_c128n25 & LDV+EX & 31.60  & 35.18 & 29.26 & 35.76 & 32.25 & 30.32 & 32.89 & 37.54 & 34.99 & 28.16 & 32.6957 & 32.0252\\
          & StageI\_c128n25\_rec6 & LDV+EX & 31.74 & 35.31 & 29.3  & 35.79 & 31.36 & 30.39 & 33.20  & 37.81 & 35.07 & 28.27 & 32.8240 & 32.1067 \\
          & StageI\_c128n25\_rec6 & LDV+cleaned\_EX & 31.81 & 35.29 & 29.29 & 35.74 & 31.4  & 30.42 & 33.34 & 37.94 & 35.11 & 28.33 & 32.8672 & 32.1934 \\
          & StageI\_c128n25\_rec6 & LDV, MSE, RMD & 31.84 & 35.35 & 29.39 & 35.78 & 31.52 & 30.45 & 33.37 & 37.97 & 35.14 & 28.36 & 32.9193 & 32.2395 \\
          & StageI\_c128n25\_rec6 & LDV, MSE, RMD, TTA & 32.09 & 35.49 & 29.48 & 35.84 & 31.62 & 30.54 & 33.60 & 38.19 & 35.38 & 28.49 & 33.0721 & 32.4334 \\
    \hline
    \multirow{5}[2]{*}{II} & SwinIR\cite{liang2021swinir} & LDV+EX & 32.05	& 35.43	& 29.40	& 35.80	& 31.57	& 30.51	& 33.61	& 38.08	& 35.23	& 28.47	& 33.0148  & 32.3687 \\
          & SwinIR\cite{liang2021swinir} & LDV+cleaned\_EX & 32.06	& 35.42	& 29.39	& 35.80	& 31.57	& 30.51	& 33.63	& 38.13	& 35.25	& 28.46	& 33.0227  & 32.3757  \\
          & SwinIR\cite{liang2021swinir} & LDV+cleaned\_EX, MSE  & 32.07 & 35.43 & 29.40 & 35.81 & 31.58 & 30.53 & 33.65 & 38.13 & 35.26 & 28.48 & 33.0327 & 32.3873 \\
          & SwinIR\cite{liang2021swinir} & \tabincell{c}{LDV+cleaned\_EX, MSE, TTA\_I}  & 32.25 & 35.54 & 29.46 & 35.86 & 31.65 & 30.59 & 33.81 & 38.28 & 35.42 & 28.58 & 33.1451 & 32.5425 \\
          & SwinIR\cite{liang2021swinir} & \tabincell{c}{LDV+cleaned\_EX, MSE, TTA\_II} & 32.27 & 35.55 & 29.47 & 35.86 & 31.66 & 30.60 & 33.83 & 38.32 & 35.45 & 28.59 & 33.1619 & 32.5525\\
    \bottomrule
    \end{tabular}%
    }
  \label{tab:tenvideo}%
\end{table*}%

\section{Experiments}
\label{sec:formatting}

\subsection{Datasets}

We use two datasets for training our models in both two stages.
First, we adopt the LDV dataset~\cite{yang2021ldv}, which is released officially by the NTIRE 2022 challenge.
It contains 240 qHD sequences belonging to 10 categories of scenes, including animal, city, closeup, fashion, human, indoor, park, scenery, sports and vehicle.
Besides, we build a large-scale dataset with 870 4K sequences acquired from YouTube.
Specially, for each above category, 87 sequences are collected.
These sequences are with high-quality and without visible artifacts.
Then, we follow the data processing procedure in NTIRE 2021 report~\cite{yang2021ntiredata}, and convert our 4K sequences to qHD sequences.
As a prepossessing, we further remove repeated frames in the compressed sequences and the corresponding frames in raw sequences.

To validate the performance of our proposed method, we select one sequence from each scenes to construct a offline validation set.
These 10 sequences are 109, 030, 125, 056, 189, 124, 119, 102, 106 and 158 from LDV dataset.
In general, we use 1100 sequences for training, and 10 sequences for validation.
\label{ref_datasets}

\subsection{Implementation Detail}

For stage I, we first fine-tune the official pre-trained BasicVSR++ model for 300K iterations with Charbonnier loss.
Adam optimizer is adopted with a initial learning rate of $2 \times 10^{-5}$.
We also adopt the Cosine Restart scheduler with the period of 300K iterations.
The learning rate is linearly increased for the first 10\% iterations.
Besides, we progressively train and converge our model by increasing the number of residual reconstruction blocks from 5 to 55.
Then, we fine-tune our model with L2 loss for 100K iterations.
All experiments are conducted with four NVIDIA V100 GPUs.

For stage II, we first fine-tune the image restoration model of SwinIR via the default Charbonnier loss, which is initialized by the pre-trained parameters on the task of image denoising.
Then we jointly fine-tune the overall model with a small learning rate of $1 \times 10^{-6}$ using L2 loss function, over our established dataset and NTIRE training dataset. 
It is noteworthy that we only sample one of every eight frames from each video for training, instead of sampling all video frames.


\subsection{Quantitative Results}

In the experiments, we adopt the peak signal-to-noise ratio (PSNR) to evaluate the video restoration performance.
We report our performance of Track 1 on two parts: (1) 10 sequences of our offline validation set and (2) 15 sequences of the official online validation set.

As shown in Table~\ref{tab:all}, our stage I model achieves 0.326 dB PSNR improvement on the offline validation set. 
Specially, by training with extra data, we improve our performance by 0.083 dB.
Besides, the performance can be further improved by 0.043 dB by removing some poor-quality sequences.
By progressively training our model, 0.128 dB PSNR improvement is achieved with 15.82M more parameters.
Fine-tuning with MSE Loss and removing duplicated frames bring us 0.028 dB and 0.014 dB improvements, respectively.

We also provide the results on our offline validation set in Table~\ref{tab:tenvideo}. As can be seen, the employing of stage II model brings 0.11 dB performance gain in terms of PSNR upon the results of stage I. Furthermore, after applying self-ensemble in stage II, the performance (\ie, PSNR) boost by 0.13 dB and achieves 33.16 dB in the offline validation set, and achieves a total improvement of 0.36 dB compared with the baseline BasicVSR++ model.
This indicates that the utilizing of stage II helps achieve superior performance, and verifies the effectiveness of our proposed two-stage strategy in restoration of compressed videos.

\subsection{Qualitative Results}

\begin{figure*}
  \centering
  \includegraphics[width=.9\linewidth]{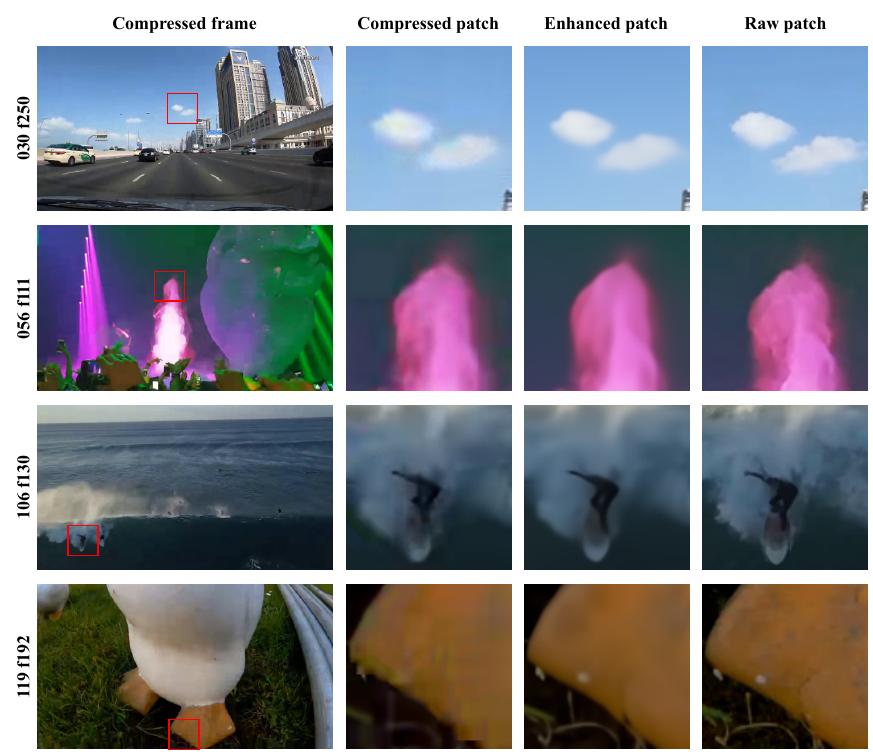}
   \caption{Subjective performance of our proposed method on our validation set.}
   \label{fig:sub2}
\end{figure*}

\begin{table*}[htbp]
  \centering
  \caption{Ablation results on the effectiveness of transfer learning for stage II in terms of PSNR $\uparrow$ and training time, respectively. The results are evaluated on our 10 offline validation videos.}
  \resizebox{0.9 \textwidth}{!}{
    \begin{tabular}{ccccccccccccc}
    \toprule
    Model & 30    & 56    & 102   & 106   & 109   & 119   & 124   & 125   & 158   & 189   & Avg.  & Training time\\
    \midrule
    SwinIR wo transfer & 31.98	& 35.39	& 29.37	& 35.79	& 31.56	& 30.46	& 33.57	& 38.09	& 35.21	&28.43	&32.9822 & 66h \\
    SwinIR wt transfer & 32.06	&35.42	&29.39	&35.80	&31.57	&30.51	&33.63	&38.13	&35.25	&28.46	& 33.0227 & 29h \\
    \bottomrule
    \end{tabular}%
    }
  \label{tab:ablation}%
\end{table*}%

We present our results on the official test set of NTIRE 2022 challenge in Fig.~\ref{fig:sub1} and those on our validation set in Fig.~\ref{fig:sub2}.
It is observed that our proposed method restore rich details in the blurred regions of video frames. Besides, the output of our solution contains less motion blur, compared with the compressed video. The edge of objects are also much clearer.

\subsection{Ablation Study}

Table \ref{tab:ablation} shows the ablation results on transfer learning of stage II. 
As can be observed, with the application of transfer learning, the PSNR of stage II achieves an improvement of 0.04 dB compared with the model training from scratch. 
This indicates the effectiveness of transferring the knowledge of image denoising to  the compressed video enhancement.
Besides, the training time of SwinIR is significantly reduced after employing transfer learning, which drops from 66 hours to 29 hours. 
This verifies the advantages of transfer learning on stage II.


\begin{table}[htbp]
  \centering
  \caption{Our results of averaged PSNR $\uparrow$ of all three tracks in the challenge. Note that for the evaluation on the validation set, we provide results of stage I/II.}
  \resizebox{0.45 \textwidth}{!}{
    \begin{tabular}{c|c|c|c}
    \toprule
    Track & \tabincell{c}{validation\\(10th frames)} & \tabincell{c}{test set\\(10th frames)} & \tabincell{c}{test set\\(all frames)} \\
    \hline
    1 (Winner) & 32.43/32.55 & 31.92 & 32.07 \\
    2 (Winner) & 28.15/28.17 & 27.48 & 27.55 \\
    3 (Runner-up) & 25.08/25.09 & 24.19 & 24.22 \\
    \bottomrule
    \end{tabular}%
    }
  \label{tab:ntire}%
\end{table}%

\section{NTIRE 2022 Challenge}
\label{sec:formatting}
We participate in all three tracks in the NTIRE 2022 super-resolution and quality enhancement of compressed video challenge.
Quantitative results are presented in Table~\ref{tab:ntire}.
In the competition, the self-ensemble~\cite{radu2015sevenways,wang2019edvr} is used in all the three tracks, while the model-ensemble is used only in Track 3.
Specifically, for Track 1\&2, we flip and rotate the input image to generate eight augmented inputs for each sample, and then merge the eight predicts as the input of the stage II model.
For Track 3, in addition to the 8 augmentation in Track 1\&2, we further conduct the model ensemble in the first stage. As a result, 16 predicts (two models with eight rotations of each) is used as the input of the stage II model.

\section{Conclusion}

In this paper, we proposed a two-stage framework to simultaneously
remove compression artifacts and mitigate the quality fluctuation in compressed videos.
Specifically, we introduced the progressive training and transfer learning strategies to stabilize the training process, shorten the training time, and improve final performance of video enhancement.
Our method achieved a good trade-off between the enhancement performance and model complexity, and wined two champions and one runner-up in the super-resolution and quality enhancement of compressed video challenge of NTIRE 2022.


\section{Acknowledgement}
This work was supported by Alibaba Group through Alibaba Research Intern Program.
{\small
\bibliographystyle{ieee_fullname}
\bibliography{ref}
}

\end{document}